\documentclass[sigconf]{aamas} 

\usepackage{balance} % for balancing columns on the final page

\usepackage{algorithm}
\usepackage{booktabs}
\usepackage{amsmath}
\usepackage{xcolor}
\usepackage{enumitem}

\usepackage[utf8]{inputenc} % allow utf-8 input
\usepackage[T1]{fontenc}    % use 8-bit T1 fonts
\usepackage{hyperref}       % hyperlinks
\usepackage{url}            % simple URL typesetting
\usepackage{booktabs}       % professional-quality tables
\usepackage{amsfonts}       % blackboard math symbols
\usepackage{nicefrac}       % compact symbols for 1/2, etc.
\usepackage{microtype}      % microtypography
\usepackage{xcolor}         % colors

\usepackage{comment}
\usepackage{amsmath}
\usepackage{algorithmicx}
\usepackage[noend]{algpseudocode}
\usepackage[dvipsnames]{xcolor}
\usepackage{subcaption}
\usepackage{graphicx}
\usepackage{wrapfig}

\usepackage{array}

\definecolor{goalAdapt_color}{HTML}{6121ef}

\definecolor{domainAdapt_color}{HTML}{f7934b}

\definecolor{inference_color}{HTML}{3ec163}

\definecolor{customgray}{rgb}{0.65, 0.65, 0.65} % Adjust the RGB values (0 to 1)
\newcommand{\ColorComment}[1]{\hfill\textcolor{customgray}{\(\triangleright\) #1}}

\newtheorem{definition}{Definition}

\usepackage{newfloat}
\usepackage{listings}
\DeclareCaptionStyle{ruled}{labelfont=normalfont,labelsep=colon,strut=off} % DO NOT CHANGE THIS
\floatstyle{ruled}
\newfloat{listing}{tb}{lst}{}
\floatname{listing}{Listing}

\setcopyright{ifaamas}
\acmConference[AAMAS '26]{Proc.\@ of the 25th International Conference
on Autonomous Agents and Multiagent Systems (AAMAS 2026)}{May 25 -- 29, 2026}
{Paphos, Cyprus}{C.~Amato, L.~Dennis, V.~Mascardi, J.~Thangarajah (eds.)}
\copyrightyear{2026}
\acmYear{2026}
\acmDOI{}
\acmPrice{}
\acmISBN{}

\acmSubmissionID{94}

\title{General Dynamic Goal Recognition using \\ Goal-Conditioned and Meta Reinforcement Learning}

\author{Osher Elhadad}
\affiliation{
  \institution{Department of Computer Science, Bar Ilan University}
  \city{Ramat Gan}
  \country{Israel}}
\email{oshereli2@gmail.com}

\author{Owen Morrissey}
\affiliation{
  \institution{Department of Computer Science, Tufts University}
  \city{Medford}
  \country{MA, USA}}
\email{morrisseyowen3@gmail.com}

\author{Reuth Mirsky}
\affiliation{
  \institution{Department of Computer Science, Tufts University}
  \city{Medford}
  \country{MA, USA}}
\email{reuth.mirsky@tufts.edu}

\begin{abstract}
Understanding an agent's goal through its behavior is a common AI problem called Goal Recognition (GR). 
This task becomes particularly challenging in dynamic environments where goals are numerous and ever-changing. 
We introduce the \textbf{General Dynamic Goal Recognition (GDGR)} problem, a broader definition of GR aimed at real-time adaptation of GR systems.  This paper presents two novel approaches to tackle GDGR: \textbf{(1)} GC-AURA, generalizing to new goals using Model-Free Goal-Conditioned Reinforcement Learning, and \textbf{(2)} Meta-AURA, adapting to novel environments with Meta-Reinforcement Learning.
We evaluate these methods across diverse environments, demonstrating their ability to achieve rapid adaptation and high GR accuracy under dynamic and noisy conditions. This work is a significant step forward in enabling GR in dynamic and unpredictable real-world environments. 

\end{abstract}


\newcommand{\BibTeX}{\rm B\kern-.05em{\sc i\kern-.025em b}\kern-.08em\TeX}

\begin{document}

%%% The following commands remove the headers in your paper. For final 
%%% papers, these will be inserted during the pagination process.

\pagestyle{fancy}
\fancyhead{}

%%% The next command prints the information defined in the preamble.

\maketitle 

%%%%%%%%%%%%%%%%%%%%%%%%%%%%%%%%%%%%%%%%%%%%%%%%%%%%%%%%%%%%%%%%%%%%%%%%

\section{Introduction}
Goal Recognition (GR) is a subfield of artificial intelligence (AI) that focuses on inferring agents' goals based on observed actions. This task is essential in a variety of fields, particularly for applications in human-robot interaction \citep{massardi2020parc,trick2019multimodal,scassellati2002theory} and multi-agent systems \citep{rabkina2019analogical,kaminka2001new,sukthankar2011activity,bansal2019beyond}, where predicting the goals and actions of other agents enables a system to respond effectively and appropriately.

\begin{figure}[t]
\begin{center}
  \includegraphics[width=0.50\textwidth]{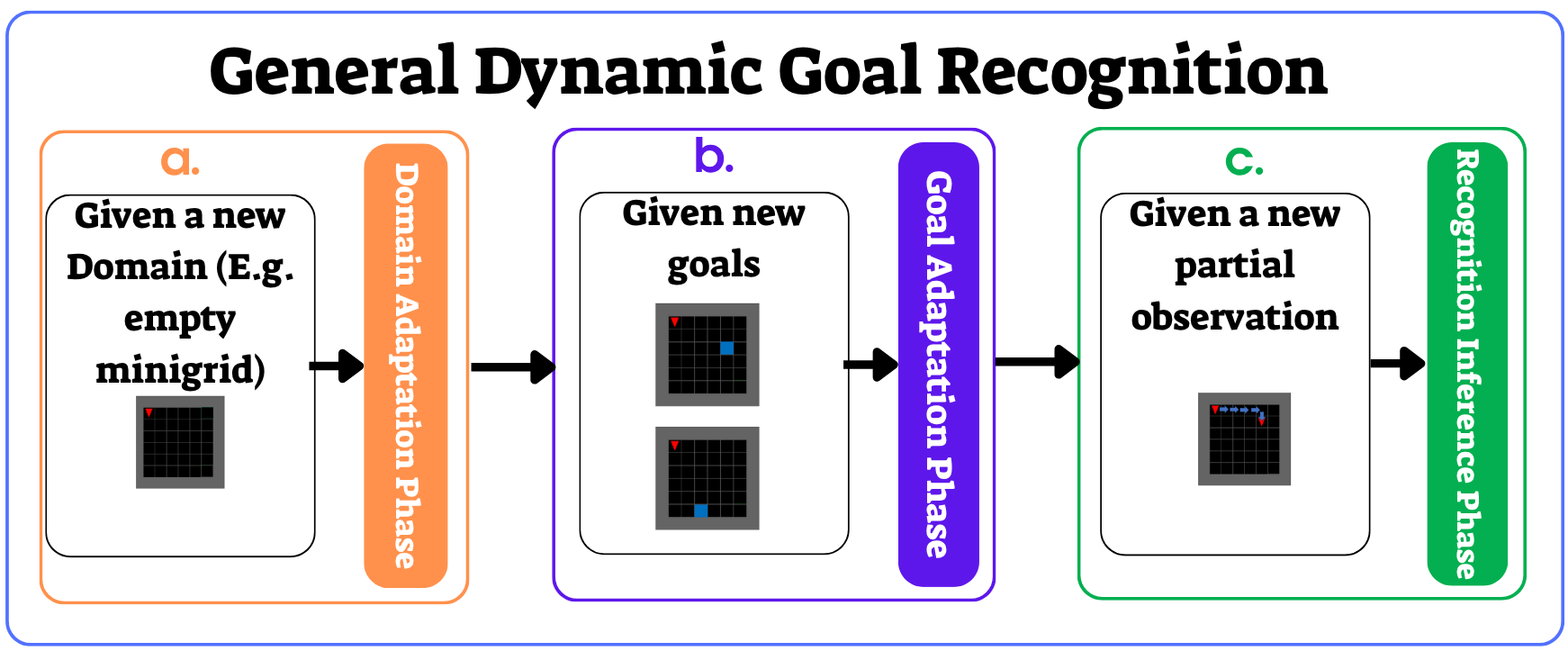}
  
  \caption{The General Dynamic Goal Recognition Problem}
  \label{real_examples}
\end{center}
\end{figure}

GR is an online inference problem, yet recent approaches rely on training models to learn patterns associated with a pre-defined set of candidate goals in a single-environment GR problem. While effective in limited settings, such approaches struggle to adapt to new sets of goals within the same or entirely novel environments, requiring them to restart their process from scratch
%While GR is an online inference problem, recent approaches leverage training to reach the inference phase with learned distributions of how likely it is for an agent to pursue each goal in a potential set of goals. Still, most existing GR solutions are primarily aimed at solving a single GR task with a specific set of goals within a single environment. When presented with a new set of goals 
 (e.g., reapplying planners or Reinforcement Learning (RL) for each new goal). This restart adds significant overhead time, rendering these approaches impractical in real-time scenarios where the goal space is continuous or there is a need for rapid adaptation to dynamically changing GR problems with diverse goals and environments. For example, in assistive technologies for the elderly or those with motor disabilities \citep{haigh2002open,zhang2017intent,massardi2020parc}, robotic assistants must adapt to changing goals across domains: inferring needed baking ingredients from gestures and context in one task, and identifying a preferred book using similar cues and prior knowledge in another. 
 % As illustrated in Figure \ref{real_examples}, on the left, the robot must infer which ingredients the elderly individual needs based on previous steps and current reaching or pointing gestures and other signals. On the right, the robot must determine the elderly individual's preferred book using a similar process involving past reading preferences and contextual observations.
 These scenarios highlight the need for an adaptive GR system, one that can generalize knowledge across tasks and transfer it to new contexts. In real-time applications, such a system cannot afford to relearn from scratch each time.
% Similarly, in autonomous vehicle systems \cite{brewitt2023verifiable}, vehicles must continuously adapt and infer the objectives of surrounding traffic agents, such as pedestrians, other vehicles, and drivers' behaviors.

This paper takes a first step towards addressing these limitations by \textbf{(1)} providing a definition for a new problem: \textbf{General Dynamic Goal Recognition (GDGR)}, a generalization of GR to account for changing goals and domains \textbf{(2)} outlining a solution paradigm for GDGR, named \textbf{Adaptive Universal Recognition Algorithm (AURA)}, and \textbf{(3)} present two implementations of AURA using Model-Free Goal-Conditioned RL (\textbf{GC-AURA}) and Meta-RL (\textbf{Meta-AURA}) to enable real-time GR across multiple dynamically changing tasks within multiple domains.

Figure~\ref{real_examples} highlights the three incremental phases that distinguish GDGR from classical GR: \textcolor{orange}{(a) \textit{Domain Adaptation}}, where the recognizer adapts to a newly introduced domain (e.g., an empty MiniGrid); \textcolor{RoyalPurple}{(b) \textit{Goal Adaptation}}, where new candidate goals are introduced within the domain; and \textcolor{green}{(c) \textit{Recognition Inference}}, where the system is required to infer the most likely goal based on partial observations. This decomposition emphasizes the dynamic nature of GDGR: unlike traditional GR, which assumes a fixed domain and goal set, GDGR requires continual adaptation across domains, goals, and observation sequences.

We present results in three environments, varying in their properties: Minigrid \citep{chevalier2024minigrid}, Point Maze \citep{gymnasium_robotics2023github}, and Panda-Gym \citep{gallouedec2021panda}. The results demonstrate that AURA significantly reduces adaptation times for new goals and environmental changes compared to existing methods, showcasing the potential of AURA to advance GR in dynamic real-world scenarios. %By addressing the run-time limitations of traditional GR systems, this research lays the groundwork for more adaptable, efficient, automated, and accurate GR applications in a wide range of changing environments.

\section{Theoretical Background}

\begin{figure*}[h]
\begin{center}
  \includegraphics[width=0.85\textwidth]{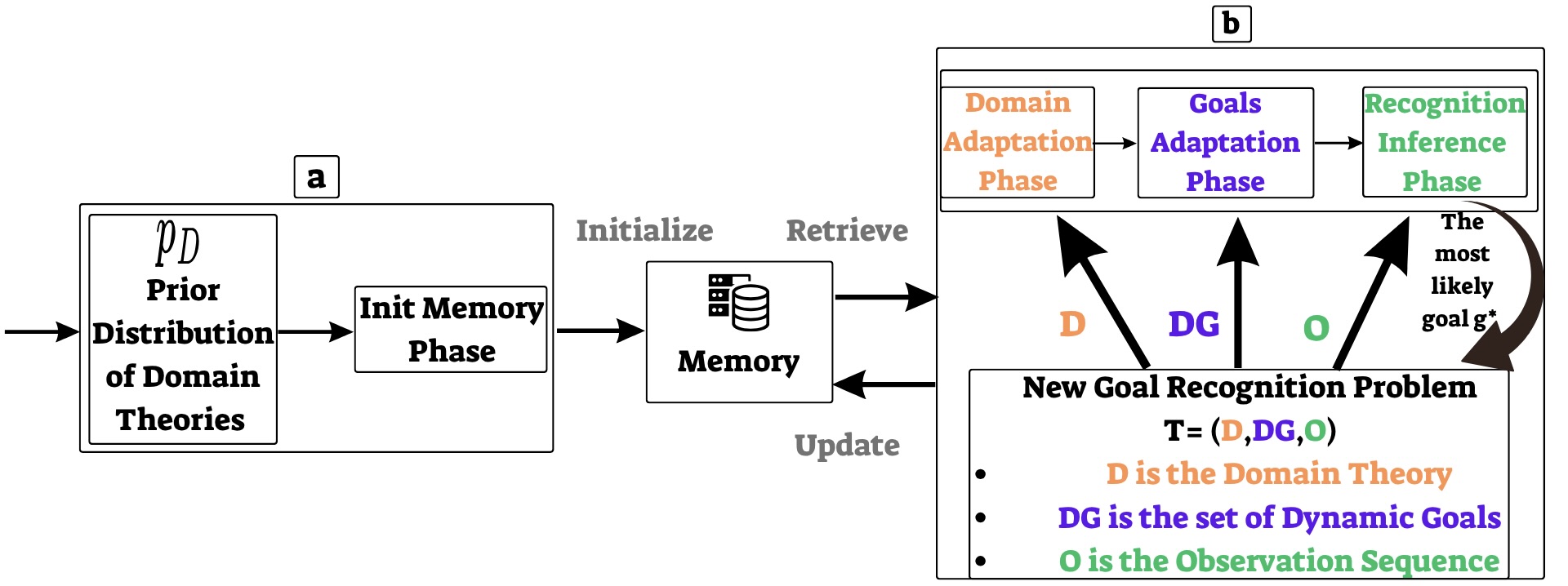}
  \caption{Adaptive Universal Recognition Algorithm (AURA).
  }
  \label{fig:DGRLL}
\end{center}
\end{figure*}

\noindent\textbf{Markov Decision Process (MDP)}  
is a mathematical framework to model an agent’s decision-making under uncertainty. It is defined as a tuple \( \mathcal{M} = (S, A, \tau, R, \gamma) \), where: \( S \) is the set of states; \( A \) is the set of actions; \( \tau: S \times A \times S \to [0, 1] \) is the state transition probability function; \( R: S \times A \to \mathbb{R} \) is the reward function; \( \gamma \in [0,1] \) is the discount factor.
Reinforcement Learning (RL) methods aim to find a policy \( \pi: S \to A \) that maximizes the expected sum of discounted rewards \citep{kaelbling1996reinforcement, sutton2018reinforcement}.

% \paragraph{Trust Region Policy Optimization (TRPO)}  
% TRPO \citep{schulman2015trust} improves policy gradient RL methods by limiting large updates, which ensures stability in optimization. The objective is to maximize the expected return \( \eta(\pi) \), subject to a constraint on the Kullback-Leibler (KL) divergence between the old and new policies:
% \begin{equation}
%     \max_\theta \mathbb{E}_{s \sim \rho_{\pi_{\text{old}}}, a \sim \pi_{\text{old}}} \left[ \frac{\pi_\theta(a|s)}{\pi_{\theta_{\text{old}}}(a|s)} A_{\theta_{\text{old}}}(s,a) \right], \quad \text{s.t.} \quad D_{\text{KL}}(\pi_{\text{old}} \,||\, \pi_\theta) \leq \delta
% \end{equation}
% Here, \( \theta \) represents the policy parameters, \( \rho_\pi \) is the state visitation frequency, and \( A_\theta(s,a) \) is the advantage function, which measures how much better action \( a \) in state \( s \) is compared to the expected action. The constraint \( D_{\text{KL}} \) limits policy changes, ensuring gradual improvement.

\noindent\textbf{Goal-conditioned Reinforcement Learning (GCRL)}  \citep{liu2022goal} adapts RL to environments where agents must achieve specific goals. A GA-MDP extends the standard MDP with an additional tuple $\langle \mathcal{G}, p_g, \phi \rangle$, where:
\( \mathcal{G} \) is the goal space; \( p_g \) is the distribution of goals; \( \phi: S \to \mathcal{G} \) maps states to goals.
The reward function depends on the state and the goal, \( R: S \times \mathcal{G} \times A \to \mathbb{R} \), and policy \( \pi \) aims to maximize the expected return for goal-directed actions:
\begin{equation}
    J(\pi) = \mathbb{E}_{a_t \sim \pi(\cdot|s_t, g), g \sim p_g, s_{t+1} \sim \tau(\cdot|s_t, a_t)} \left[\sum_{t} \gamma^t R(s_t, a_t, g)\right]
\end{equation}

\noindent\textbf{Meta-Reinforcement Learning (Meta-RL)}  
 is a problem formulation, which aims to train agents that can quickly adapt to new tasks by leveraging experience from a distribution of related tasks. Formally, consider a distribution over tasks \( p(\mathcal{T}) \), where each task \( \mathcal{T}_i \) is modeled as an MDP. The objective is to learn a policy \( \pi_\theta(a \mid s) \), parameterized by \( \theta \), that can be efficiently adapted to any task \( \mathcal{T}_i \) with minimal data. Concretely, we assume all tasks share the same state and action spaces and policy parameterization, but may differ in their transition dynamics \(\tau\) or reward functions \(R\). Meta-RL methods learn a common initialization of policy parameters \(\theta\) that -- after a small number of additional gradient steps -- yields strong performance on any new task drawn from this distribution.
A canonical algorithm is Model-Agnostic Meta-Learning (MAML) \citep{finn2017model}.% , which alternates between two nested loops over tasks:

%\begin{itemize}
  %\item \emph{Inner loop:} For each task \( \mathcal{T}_i \), adapt the initial parameters \( \theta \) to task-specific parameters \( \theta'_i \) using one or more gradient descent steps on the task's loss:
  %\[
    %\theta'_i = \theta - \alpha \nabla_\theta \mathcal{L}_{\mathcal{T}_i}(\theta),
  %\]
  %where \( \alpha \) is the inner-loop learning rate, and \( \mathcal{L}_{\mathcal{T}_i}(\theta) \) denotes the loss function for task \( \mathcal{T}_i \), typically defined as the negative expected return:
  %\[
    %\mathcal{L}_{\mathcal{T}_i}(\theta) = -\mathbb{E}_{\tau \sim \pi_\theta} \left[ \sum_{t=0}^H \gamma^t R_i(s_t, a_t) \right],
  %\]
  %with \( \tau = (s_0, a_0, \ldots, s_H, a_H) \) representing a trajectory sampled under policy \( \pi_\theta \).
  
  %\item \emph{Outer loop:} Update the initial parameters \( \theta \) to minimize the loss across tasks after adaptation:
  %\[
    %\min_{\theta} \mathbb{E}_{\mathcal{T}_i \sim p(\mathcal{T})} \left[ \mathcal{L}_{\mathcal{T}_i}(\theta'_i) \right].
  %\]
%\end{itemize}

%This bi-level optimization trains the model to be sensitive to task-specific updates, enabling efficient adaptation to new tasks with limited data.

\noindent\textbf{Goal Recognition (GR)} is the task of inferring the likely goal of an observed % \ben{Did you mean observed?}
agent according to a series of observations \citep{sukthankar2011activity,meneguzzi2021survey}. %If accepted, add mirsky, keren and Geib's survey
\begin{definition}
    A GR problem is defined as a tuple \( T = (D, G, O) \), where \( D \) is the domain theory, \( G \) is the set of potential goals, and \( O \) is a sequence of observations. The objective of GR is to identify a goal \( g \in G \) that best explains the observation sequence \( O \).
\end{definition}
Different approaches to GR primarily differ in how they formulate the domain theory \( D \), including planning domains \citep{ramirez2009plan}, grammars \citep{geib2018learning}, process graphs \citep{ko2023plan}, or GA-MDPs \citep{ramirez2011goal, amado2022goal}.

While GR focuses on identifying the specific goal an agent is pursuing based on a partial execution of its behavior, Inverse Reinforcement Learning (IRL) \citep{arora2021survey} seeks to recover the underlying reward function that explains the agent’s behavior across multiple trajectories. GR is typically formulated as an online, one-shot inference task, often under partial observability, where the aim is to determine the most likely goal from a limited observation sequence. In contrast, IRL assumes access to multiple trajectories and aims to generalize the agent’s preferences. % through a learned reward function.

% \paragraph{Goal Recognition as Reinforcement Learning}
% \citet{amado2022goal} define a GR as RL problem as: \matan{Unless you're using this definition to GR explicitly in this paper, in the light of the last statement before this definition, why is it worth mentioning/defining? there are other formulations for GR that aren't used here and aren't mentioned in this section.}
 % \( T = (D, G, O) \) where:  \( G \) - the set of potential goals;  \( O \) - a sequence of observations; \( D \) - the domain theory, which is represented as a set of policies, one per goal \(D= \{\pi_g | g \in G\} \).

% divided into two types:
        % \begin{itemize}
        %     \item Utility-based domain theory \( D_Q(G) \), represented as \( (S, A, Q) \) where \( Q \) is a set of Q-functions \( \{Q_g | g \in G\} \).
        %     \item Policy-based domain theory \( D_\pi(G) \), represented as \( (S, A, \pi) \) where \( \pi \) is a set of policies \( \{\pi_g | g \in G\} \).
        % \end{itemize}

\section{The General Dynamic Goal Recognition (GDGR) Problem}

%\reuth{Start with a motivating example for changing goals. Explain why existing work can't handle it and then show GDGR.}
% We introduce a generalization to the GR problem as a continuous transfer learning task \matan{I'd remove the 'continuous', seems unrelated to the term, but rather to the environments supported.}, referred to as General Dynamic Goal Recognition (GDGR).
We introduce a transfer‐learning formulation of the GR problem - General Dynamic Goal Recognition (GDGR) - which handles sequences of GR problems with changing goals and domains. Intuitively, GDGR consists of a series of GR problems, where each problem may involve a new observation sequence and potentially introduce previously unseen goals or changes in the underlying domain. The colors throughout the paper match the colors in Figure \ref{fig:DGRLL} to provide visual assistance in decomposing GDGR into single recognition tasks.

\begin{definition}
\label{def:gdgr} \textbf{General Dynamic Goal Recognition (GDGR)} is
a tuple representing the prior distribution of domain theories \( p_D \) and a sequence of GR problems:
\[
\langle p_D, (T_1 = \langle \textcolor{orange}{D}_1, \textcolor{RoyalPurple}{DG}_1, \textcolor{green}{O}_1 \rangle, T_2, \ldots, T_n) \rangle,
\]
where each input is provided at an increasing time step, starting with \( p_D \) at time step \( t = 0 \), and each GR problem corresponds to a distinct time step \( t \in \{1, \ldots, n\} \). 
\end{definition}

For a given series of time steps \( t \in \{1, \ldots, n\} \), each time step consists of three stages of inputs, which can be given incrementally: The first input is the domain theory \( \textcolor{orange}{D}_t \); The second input is the set of dynamic goals used for recognition \( \textcolor{RoyalPurple}{DG}_t\); and the last input is an observation sequence 
    $
    \textcolor{green}{O}_t.
    $
     % that can have gaps between states and actions. %\osher{Does every GR is an online problem?} Note: as GR is an online setting where state and action pairs appear incrementally. % and conform with the order in which they appear within the sequence.
% \end{itemize}

\begin{algorithm*}[h]

\caption{Adaptive Universal Recognition Algorithm (AURA)}
\begin{algorithmic}[1]
\Require{$p_D$ - prior distribution of domain theories}
\State Init Memory $M$
\State $M \gets \text{InitMemoryPhase}(p_D, M)$ \ColorComment{Initialized $M$ after adaptation to $p_D$}
\For{all $T_i$ in $\text{GetGoalRecognitionProblem}()$}
\State Get Domain theory $\textcolor{orange}{D}_i$ from $T_i$
\State $M_{\textcolor{orange}{D}_i} \gets \text{DomainAdaptationPhase}(\textcolor{orange}{D}_i, M)$ \ColorComment{Domain Memory $M_{\textcolor{orange}{D}_i}$ after domain adaptation}
\State Get the set of new dynamic goals $\textcolor{RoyalPurple}{DG}_i$ from $T_i$
\State $\{M_g\}_{g\in \textcolor{RoyalPurple}{DG}_i} \gets \text{GoalsAdaptationPhase}(\textcolor{RoyalPurple}{\textcolor{RoyalPurple}{DG}_i}, M_{\textcolor{orange}{D}_i})$ \ColorComment{Goals Memory $\{M_g\}_{g\in \textcolor{RoyalPurple}{DG}_i}$ after goal adaptation}
\State Get the Observation sequence $\textcolor{green}{O}_i$ from $T_i$ \ColorComment{In the online GR setting, each state and action tuple is provided in a different time step and has its own recognition inference phase}
\State $g^* \gets \text{RecognitionInferencePhase}(\{M_g\}_{g\in \textcolor{RoyalPurple}{DG}_i}, \textcolor{green}{O_i})$ \ColorComment{$g^* \gets \text{arg max}_{g\in \textcolor{RoyalPurple}{DG}_i}[\text{DISTANCE}(\textcolor{green}{O}_i, M_g)]$, where DISTANCE calculates the similarity between the observation sequence and the Goals Memory}
\State Save and return $g^*$
\State $M \gets \text{UpdateMemoryPhase}(M, M_{\textcolor{orange}{D}_i}, \{M_g\}_{g\in \textcolor{RoyalPurple}{DG}_i}, T_i, g^*)$ \ColorComment{Updated $M$ using the previous $M$, and the current GR problem, adaptations and inference}
\EndFor
\end{algorithmic}
\label{alg:GDGR}
\end{algorithm*}

This work focuses on GA-MDPs as the domain theory, specifically \( \textcolor{orange}{D} = (S, A, \tau) \), where \( S \) is the set of states, \( A \) is the set of actions, and \( \tau \) is the state transition probability.
%
%The definitions of \( G \) and \( O \) also vary across GR approaches. 
In GA-MDP-based formulations of GR, the observation sequence \( \textcolor{green}{O} \) is typically defined as $\textcolor{green}{O} = (\langle s_{1}, a_{1} \rangle, \langle s_{2}, a_{2} \rangle, \ldots)
$, a sequence of state-action pairs, which may be consecutive or non-consecutive, where each pair denotes a state-action transition observed over time. The goal set \( \textcolor{RoyalPurple}{DG} \subseteq \mathcal{G} \) is assumed to be a subset of the broader goal space \( \mathcal{G} \) defined by the GA-MDP.

The solution of a GDGR problem $\langle p_D,  (T_1, \ldots, T_n) \rangle$ is a sequence of goals \( (g^*_1, \ldots, g^*_n) \) for each of the single GR problems \( t \in \{ 1, \ldots, n \} \), such that
$g_t^* = \arg\max_{g \in \textcolor{RoyalPurple}{DG}_t} P(g|\textcolor{green}{O}_t, M_{t-1})$. \( g^*_t \) is the recognized goal within the set of dynamic goals \( \textcolor{RoyalPurple}{DG}_t \) based on the given observations \( \textcolor{green}{O}_t \) and the \textit{Memory} \( M_{t-1} \) carried over from the previous time step.  Memory refers to the structured system or set of mechanisms responsible for storing, maintaining, and retrieving information that helps to transfer knowledge between iterations.

\begin{table*}[H]
\centering
\renewcommand{\arraystretch}{1.2} % slightly increase row height for readability
\begin{tabular}{m{2cm} m{2cm} m{9cm}}
    \toprule
    \textbf{Figure} & \textbf{Domain} & \textbf{Details} \\
    \midrule
    \includegraphics[width=1.5cm]{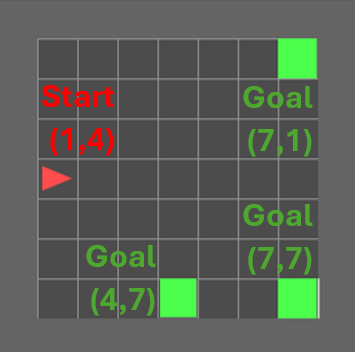} 
    & \raggedright\arraybackslash \textbf{MiniGrid} 
    & \raggedright\arraybackslash \textit{Environments}: Empty-MiniGrid-9x9 environment. \newline \textit{Motivation}: Discrete navigation; comparison with GR baselines. \newline
      \textit{States}: Image + direction of agent (Discrete). \newline \textit{Actions}: Turn, move forward, stay in place (Discrete). \newline
      \textit{Reward}: $1 - 0.9 * (step_{count} / max_{steps})$ for success or 0 for failure (Sparse). \\
    \midrule
    \includegraphics[width=1.5cm]{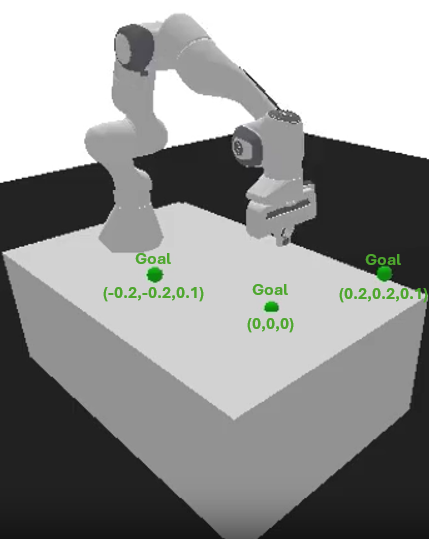} 
    & \raggedright\arraybackslash \textbf{Panda-Gym} 
    & \raggedright\arraybackslash \textit{Environments}: PandaReach environment. \newline \textit{Motivation}: Realistic 3D robotic control scenario. \newline
      \textit{States}: Robotic arm positions, velocities (Continuous). \newline \textit{Actions}: Joint torques for robotic arm (Continuous). \newline
      \textit{Reward}: Gradual reward based on distance to goal (Dense). \\
    \midrule
    \includegraphics[width=1.5cm]{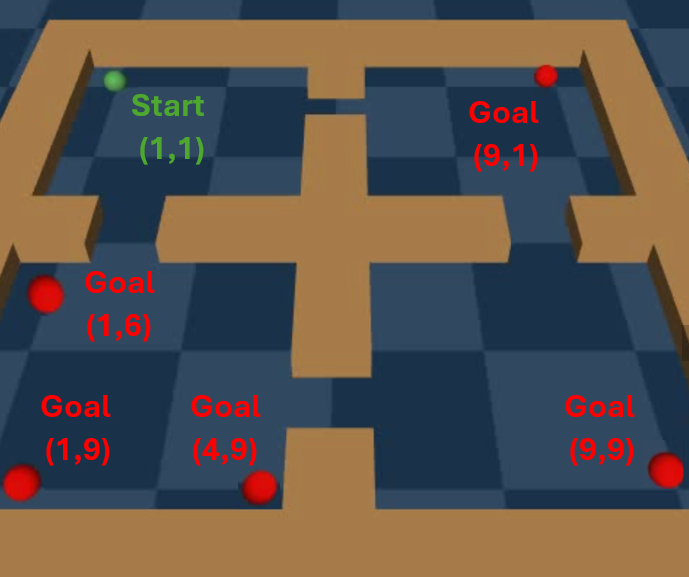} 
    & \raggedright\arraybackslash \textbf{PointMaze} 
    & \raggedright\arraybackslash \textit{Environments}: 4-Rooms-11x11 environment. \newline \textit{Motivation}: Continuous navigation; path complexity. \newline
      \textit{States}: Agent positions, velocities (Continuous). \newline \textit{Actions}: Force applied to agent (Continuous). \newline
      \textit{Reward}: -1 if not reaching goal; 0 when near goal ($<$ 0.5m) (Sparse). \\
    \bottomrule
\end{tabular}
\caption{Comparison of domains, visualizations, and their specific characteristics. Each domain is chosen to highlight different capabilities of the AURA framework.} % across discrete and continuous spaces.}
\label{tab:domains}
\end{table*}

\section{Adaptive Universal Recognition Algorithm (AURA)}

To solve GDGR problems, we introduce the \textit{Adaptive Universal Recognition Algorithm (AURA)} -- a generic algorithm that leverages the fact that domain theories, goal specifications, and online observations may arrive at different times. AURA operates in three main phases. First, it initializes a global \textit{Memory}, which can later reduce the computational cost of adapting to new problems. Then, for each \textit{GR problem}, AURA performs task-specific \textit{domain and goal adaptation} to tailor its knowledge to the new setting. Finally, once an observation sequence is received, AURA infers the most likely goal and updates the memory. %, enabling continual learning through knowledge transfer.
Figure~\ref{fig:DGRLL} and Algorithm~\ref{alg:GDGR} outline AURA's core components:

\noindent \textbf{1. Initial Memory Phase} (Algorithm~\ref{alg:GDGR}, line~2 -- \texttt{InitMemoryPhase}; Figure~\ref{fig:DGRLL}, component (a)) 
Given the prior distribution \( P_D \), this phase initializes the memory before any GR problem is received. Its objective is to pre-process reusable domain or goal-related knowledge to reduce adaptation and inference costs in later phases.

\noindent \textbf{2. GR Problem Processing} (Algorithm~\ref{alg:GDGR}, lines~3–10; Figure~\ref{fig:DGRLL}, component (b)) 
For each GR problem, AURA processes the domain theory, dynamic goals, and observation sequence, which may arrive at different times.
\begin{itemize}[leftmargin=*, itemsep=1pt, topsep=0pt]
    \item \textbf{Domain Adaptation Phase:} 
    Upon receiving the domain theory \textcolor{orange}{D}, AURA applies the \texttt{DomainAdaptationPhase} to retrieve or refine domain-specific knowledge from memory. This adaptation supports subsequent goal reasoning and recognition.

    \item\textbf{Goal Adaptation Phase:} 
    When the dynamic goals \textcolor{RoyalPurple}{DG} become available, AURA invokes the \texttt{GoalsAdaptationPhase} to tailor internal representations to the specific goal set. This enables efficient and accurate recognition later.

    \item \textbf{Recognition Inference Phase:} 
    Once the sequence \textcolor{green}{O} is available, AURA performs inference using the \texttt{RecognitionInferencePhase} to identify the most likely goal. % based on the adapted knowledge.
\end{itemize}

\noindent \textit{Note:} The components of a GR problem -- domain, goals, and observations -- are not assumed to arrive simultaneously. AURA is designed to conceptually take advantage of any time gaps between their arrivals. This enables early adaptation steps, such as beginning domain or goal learning before observations are received.

\noindent \textbf{3. Memory Update Phase} (Algorithm~\ref{alg:GDGR}, line~11 -- \texttt{UpdateMemoryPhase}) \\
After solving each GR problem, AURA updates its memory with the new results and learned patterns. This formulation supports continual learning and improves future performance by transferring accumulated knowledge across problems.

%\subsection{GDGR Abstraction Levels}
\noindent\textbf{AURA Abstraction Levels} %\reuth{generalizability?}
Algorithm \ref{alg:GDGR} is a generic algorithm. It enables generalization at various levels of the problem, from basic GR problems with new observations, through changing goals, or new domain dynamics. AURA can be implemented in various ways to address each type of these \textit{abstractions} that can be classified into three levels:
\textbf{(1)} GR problems with a fixed set of goals within a single domain (\textcolor{green}{O} is changing between time steps, but \textcolor{orange}{D} and \textcolor{RoyalPurple}{DG} are fixed). \textbf{(2)} GR problems with many possible sets of goals within a single domain (\textcolor{green}{O} and \textcolor{RoyalPurple}{DG} are changing between time steps, while \textcolor{orange}{D} is fixed). \textbf{(3)} GR problems with many possible goals across multiple domains (changing \textcolor{green}{O}, \textcolor{RoyalPurple}{DG} and \textcolor{orange}{D}).

Next, we introduce two RL-based implementations of AURA, each targeting a different level of abstraction within the framework. The first, GC-AURA, builds on GCRL to enable fast adaptation to a new set of goals, addressing AURA’s second level of abstraction. The second, Meta-AURA, is based on Meta-RL and targets AURA’s third level of abstraction by enabling adaptation to new domains that share the same state and action spaces and can vary in their transition and reward functions.

\subsection{GC-AURA}

This approach targets GDGR problems in which all GR problems share the same domain theory \( \textcolor{orange}{D} \). In this setting, the memory \( M \) consists of a GCRL policy that can be trained over the entire goal space of \( \textcolor{orange}{D} \) or on a subset of the goal space. This enables AURA to reuse prior domain knowledge for new recognition problems without retraining from scratch.

The primary challenge here is to adapt this policy to newly introduced goals \( \textcolor{RoyalPurple}{DG} \) with minimal additional training. This requires a mechanism for lifelong goal adaptation, where the system continuously improves its ability to generalize across goals within the same domain.

During the \texttt{DomainAdaptationPhase}, a GCRL policy is trained on the domain goal space, or a goal subspace. This shared policy serves as a reusable basis for all subsequent recognition problems in that domain. Once a new set of goals is introduced, the \texttt{GoalsAdaptationPhase} uses transfer learning, for example, can use the following strategies:

\begin{enumerate}[leftmargin=*, itemsep=2pt, topsep=2pt]
    \item \textbf{Zero-shot transfer:} Directly use the existing GCRL policy without any additional training.
    \item \textbf{Few-shot adaptation:} Fine-tune the policy on a specific goal using a small number of gradient steps or episodes. Useful when the goal is substantially different from those seen in earlier steps.
    \item \textbf{Goal recall:} Retrieve a previously stored goal-specific policy if the exact goal has been encountered in a past GR problem.
\end{enumerate}

Zero-shot and recall are generally preferred for their efficiency. However, some goals may require minor updates of the policy to ensure accurate recognition. Few-shot adaptation is motivated by the need to handle goals that lie in underrepresented regions of the goal space or whose reward dynamics significantly differ from previously seen goals. %This provides a trade-off between recognition speed and accuracy, depending on the task constraints.

In Section~\ref{sec:gc_aura_exp}, we demonstrate GC-AURA using TRPO \citep{schulman2015trust} to train a GCRL policy over a continuous goal space, where zero-shot transfer was found to be effective in the Panda-Gym domain.

\subsection{Meta-AURA}
In RL-based GR approaches, new domains can differ in all MDP characteristics. For our Meta-AURA implementation using MAML-TRPO, the state and action spaces are assumed to be the same, and the transitions and rewards can change between different domains.
This approach focuses on cases where GR problems span across multiple domain theories \(\textcolor{orange}{D}\). The Memory $M$ is represented by a meta-policy trained to generalize over diverse domains with varying transitions and rewards. This meta-policy %is designed to 
improves adaptation times to novel goals and domains by leveraging prior experience.

The primary challenge is to enable the meta-policy to efficiently adapt to a new domain \(\textcolor{orange}{D}\) and its associated dynamic goals \(\textcolor{RoyalPurple}{DG}\). This process demands a robust initialization strategy and domain adaptation mechanism to minimize adaptation time while maintaining high recognition accuracy.

During the \texttt{InitMemoryPhase} we train a meta-policy for all domains from the prior distributions $p_D$. The \texttt{GoalsAdaptationPhase} evaluates each goal in the dynamic goal set \(\textcolor{RoyalPurple}{DG}\) and decides between: (1) Few-shot adaptation of the meta-policy to specific goals, (2) Few-shot adaptation of the meta-policy to GCRL policy over all goals in \(\textcolor{RoyalPurple}{DG}\), or (3) Retrieval of previously adapted goal-specific or GC policies for recurrent goals from earlier GR problems. 

In Sections \ref{sec:meta_discrete} and \ref{sec:meta_continuous}, we leveraged the MAML-TRPO algorithm to train a meta-policy capable of adapting to a diverse set of domains and goals. This meta-policy was later fine-tuned using TRPO to specific goals and domains.

\section{Experimental Setup}
\label{sec:experiments}

We start by detailing the general setup configurations used across all experiments.

\begin{table*}[t]
\begin{tabular}{m{2cm} m{2cm} m{9cm}}
    \toprule
    \textbf{Figure} & \textbf{Domain} & \textbf{Details} \\
    \midrule
    \includegraphics[width=1.5cm]{Figures/grid3.PNG} 
    & \raggedright\arraybackslash \textbf{MiniGrid} 
    & \raggedright\arraybackslash \textit{Environments}: Empty-MiniGrid-9x9 environment. \newline \textit{Motivation}: Discrete navigation; comparison with GR baselines. \newline
      \textit{States}: Image + direction of agent (Discrete). \newline \textit{Actions}: Turn, move forward, stay in place (Discrete). \newline
      \textit{Reward}: $1 - 0.9 * (step_{count} / max_{steps})$ for success or 0 for failure (Sparse). \\
    \midrule
    \includegraphics[width=1.5cm]{Figures/panda_gym_goals.png} 
    & \raggedright\arraybackslash \textbf{Panda-Gym} 
    & \raggedright\arraybackslash \textit{Environments}: PandaReach environment. \newline \textit{Motivation}: Realistic 3D robotic control scenario. \newline
      \textit{States}: Robotic arm positions, velocities (Continuous). \newline \textit{Actions}: Joint torques for robotic arm (Continuous). \newline
      \textit{Reward}: Gradual reward based on distance to goal (Dense). \\
    \midrule
    \includegraphics[width=1.5cm]{Figures/point_maze_4_rooms_1x9_9x1_9x9_1x6_4x9.png} 
    & \raggedright\arraybackslash \textbf{PointMaze} 
    & \raggedright\arraybackslash \textit{Environments}: 4-Rooms-11x11 environment. \newline \textit{Motivation}: Continuous navigation; path complexity. \newline
      \textit{States}: Agent positions, velocities (Continuous). \newline \textit{Actions}: Force applied to agent (Continuous). \newline
      \textit{Reward}: -1 if not reaching goal; 0 when near goal ($<$ 0.5m) (Sparse). \\
    \bottomrule
\end{tabular}
\caption{Comparison of domains, visualizations, and their specific characteristics. Each domain is chosen to highlight different capabilities of the AURA framework.} % across discrete and continuous spaces.}
\label{tab:domains}
\end{table*}

\subsection{Domains}
\label{app:domains}

AURA supports both discrete and continuous environments, making it suitable for evaluation under a variety of conditions. We evaluate its variants across three benchmark domains from OpenAI Gymnasium \citep{towers2024gymnasium}: MiniGrid, Point-Maze, and Panda-Gym. Table~\ref{tab:domains} summarizes the key differences between these domains.

\paragraph{MiniGrid}
\label{sec:minigrid}

The MiniGrid domain was chosen due to its flexibility in representing environments in multiple formats, which makes it well-suited for modeling complex GR problems. This capability allows us to compare diverse approaches under consistent conditions, thereby offering insights into algorithmic performance across different representations. Our study focuses MiniGrid Empty 9x9 environment (shown in Table \ref{tab:domains}). Specifically, we compared AURA to three baseline algorithms: PDDL for reasoning and GR (R\&G) using a complete environment model, GRAQL with symbolic state representations (coordinates and angle), and DRACO with a visual image-based representation (same as AURA visual image-based representation). Each representation provides the algorithms with varying degrees of information, enabling a comprehensive comparison of their strengths and limitations when tackling the same GR problem. We used Minigrid Empty 9x9 with different goal locations and lava locations as the training environments for MAML training. The adaptation rewards for MAML in this domain are shown in the Appendix, in Figure \ref{fig:minigrid_maml_adaptation_rewards}.

\paragraph{Point-Maze}
\label{sec:pointmaze}

The Point-Maze domain presents a continuous environment where the agent navigates through a maze, starting from a fixed position and aiming to reach one of three predefined goal locations. This domain is particularly valuable for testing algorithms in continuous state, goal, and action spaces, as it introduces challenges such as path complexity, exploration efficiency, and obstacle avoidance. The environment's structure and continuous nature make it a suitable testbed for evaluating Meta-AURA. By assessing performance in this domain, we gain insights into the adaptability and effectiveness of different approaches in navigating and achieving goals within complex spatial constraints. We used Mazes with different sizes: 6x6 - 9x9, and different start goal states, and obstacles locations as the training environments for MAML training. The adaptation rewards for MAML are shown in Figure \ref{fig:point_maze_maml_adaptation_rewards}. %Fine-tuning the MAML policy on 4-Rooms 11x11 environment is reported in Table \ref{tab:domains}.

\paragraph{Panda-Gym} 
\label{sec:pandagym}

The Panda-Gym domain features a robotic arm tasked with manipulating objects to reach predefined goal positions within a continuous 3D space. This environment poses unique challenges by combining high-dimensional state and action spaces. The robotic arm begins at a default position and must plan and execute precise movements to achieve one of three target goals. Panda-Gym provides a realistic simulation for evaluating reinforcement learning algorithms, especially in robotics applications, as it tests spatial reasoning, motion control, and goal-oriented behavior. This domain highlights the robustness of algorithms when applied to real-world-inspired scenarios involving intricate control dynamics and the need for precision.

%generated GR problems for each domain by creating diverse configurations with shared initial setups and strategically placed obstacles.

\subsection{Baselines.}
(1) \textit{Plan Recognition as Planning (R\&G):} This classic symbolic GR baseline, proposed by \citet{ramirez2009plan}, leverages planner executions to infer the likelihood of goals. 
% \cite{ramirez2011goal}
 It relies on an underlying PDDL domain representation, meaning
%, which can either be manually crafted \cite{vered2017online} or derived from labeled data \cite{granada2020object}. 
 without significant and non-trivial modifications, it is limited to discrete domains, making it unsuitable for continuous setups. Consequently, we use it exclusively for evaluation in MiniGrid  (Section \ref{sec:meta_discrete}). % by referring to the results reported in Nageris \textit{et al}. \shortcite{nageris2024goal}. 

(2) \textit{GRAQL:} A Q-learning-based approach that learns a tabular policy for each goal and infers the most likely goal based on these \citep{amado2022goal}. %to compute the likelihood of each goal hypothesis. 
As GRAQL is inherently tabular, it is also restricted to discrete domains. Thus, as with R\&G, we compare it to AURA only on the MiniGrid domain (Section \ref{sec:meta_discrete}).  %, using the results reported in Nageris et al. \shortcite{nageris2024goal}.

(3) \textit{DRACO:} GR method that supports discrete and continuous states, goals, and action spaces using deep RL \citep{nageris2024goal}. In the continuous domains: Panda-Gym and Point-Maze (Sections \ref{sec:gc_aura_exp} and \ref{sec:meta_continuous} respectively),
our Goal-Conditioned and Meta-RL algorithms are based on TRPO, while DRACO's original implementation uses PPO. To better isolate recognition performance and ensure a fair comparison with our approach, we modified DRACO to use TRPO as its underlying RL algorithm. This change avoids confounding effects from differences in RL strategy. %In different settings, DRACO training was referred to as Goal-Directed RL policy trained from scratch (compared to GCRL with Goal-Conditioned RL or Meta-RL with Goal-Directed RL policy trained from Meta-policy).

% we use Trust Region Policy Optimization (TRPO) instead of the algorithm's original implementation using PPO, as our goal-conditioned policies and meta-RL algorithms are based on TRPO. We made this modification so all algorithms use the same underlying RL algorithm.

\subsection{Policy Learning Algorithms.}
We implemented TRPO and MAML-TRPO using the Learn2Learn Library \citep{Arnold2020-ss}. 
% \reuth{Discuss parallel being orthogonal to learning.}
%MAML enables efficient adaptation to new tasks by learning a meta-policy that can quickly fine-tune to new goals \cite{finn2017model}. TRPO, a policy gradient method, optimizes policies while maintaining trust regions to ensure stable and reliable updates, making it particularly suitable for goal-conditioned tasks in continuous domains.
 All information on computational resources and efficiency is provided in Appendix~\ref{sec:compute}, information on hyperparameters and training details are presented in Appendix~\ref{appendix:hyperparameters},, and details about licenses and external packages can be found in Appendix~\ref{sec:licenses}.

\subsection{GR Methods.} We evaluate recognition performance, framed as a classification task, using standard metrics: Accuracy, Recall, Precision, and F-score. In addition, we employ two distinct distance metrics between observed trajectories and RL policies that are used by the GR algorithm for recognition inference (\texttt{DISTANCE} in Algorithm \ref{alg:GDGR}). In discrete state spaces (Section \ref{sec:meta_discrete}, Algorithms: Meta-AURA, DRACO, and GRAQL), we implement the recognition inference using Kullback–Leibler (KL) divergence as the recognition metric:
\begin{equation}
    D_{\text{KL}}(\pi_g \| \pi_O) = \sum_{i \in |O|} \pi_g(a_i|s_i) \log \frac{\pi_g(a_i|s_i)}{\pi_O(a_i|s_i)}
\end{equation}
where $\pi_g$ is a goal-dependent softmax policy and $\pi_O$ represents a pseudo-policy derived from observations $O$. For continuous goal spaces (Sections: \ref{sec:gc_aura_exp} and \ref{sec:meta_continuous}, Algorithms: GC-AURA, Meta-AURA, and DRACO), we utilize the Wasserstein distance metric, as introduced in Nageris et al. for recognition purposes \citep{nageris2024goal}:
\begin{equation}
    W(O, \pi) = \mathbb{E}_{(s,a) \in O}[\|a - \tilde{a}\|_{L_1}], \quad \tilde{a} \sim \pi(s)
\end{equation}
where $W(O, \pi)$ is the Wasserstein distance between observation sequence $O$ and policy $\pi$; $\mathbb{E}{(s,a) \in O}$ is the expected value over state-action pairs in observation sequence; $|a - \tilde{a}|_{L_1}$ is the L1 norm (absolute difference) between observed action $a$ and sampled action $\tilde{a}$; $\tilde{a} \sim \pi(s)$ is an action $\tilde{a}$ sampled from policy $\pi$ given state $s$.

\begin{figure}[t]
\centering
\begin{minipage}[t]{0.48\textwidth}
    \centering
    \includegraphics[width=\textwidth]{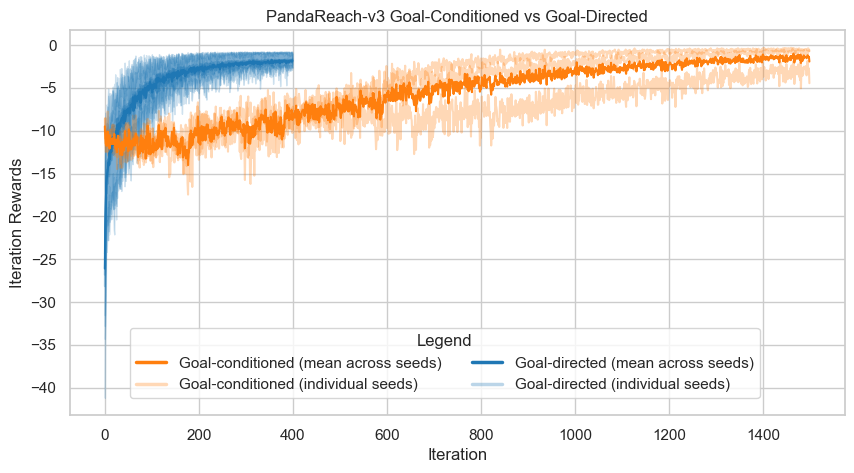}
    \caption{Training curves for the PandaReach-v3 domain. The \textbf{orange curve} represents the Goal-Conditioned TRPO policy (for GC-AURA) trained across all goals in the continuous goal space. 
    % This policy, while slower to train, allows for immediate adaptation to any goal without requiring additional training \reuth{this sentence should be in the main text, not the caption}. 
    The \textbf{blue curve} shows the goal-directed TRPO policy (for DRACO) trained for a specific goal. The shaded regions indicate variations across changing seeds and goals.}
    \label{fig:gc_aura_train}
\end{minipage}
\hfill
\begin{minipage}[t]{0.48\textwidth}
    \centering
    \includegraphics[width=\textwidth]{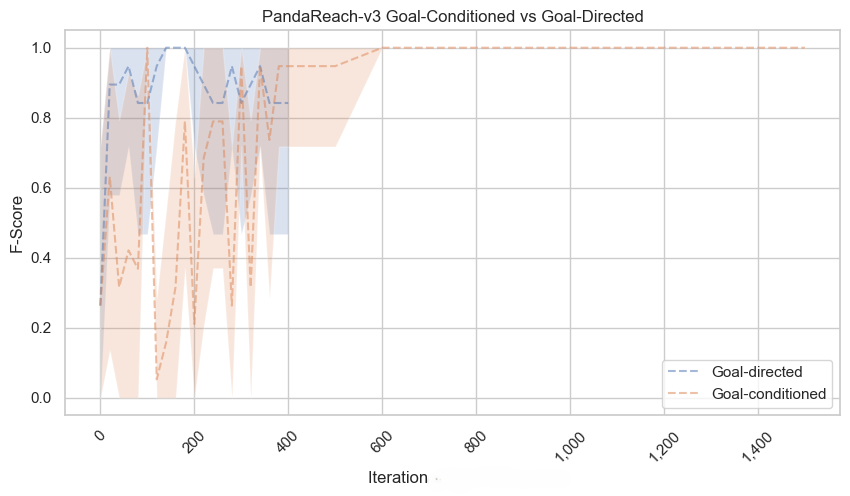}
%     \caption{F-Score for PandaReach environment with 10\% observability rate and 0-90\% noise rate, evaluated using the Wasserstein GR distance metric across 20 different GR tasks. The \textbf{orange curve} represents the F-Score of the GC-TRPO policy across different iterations. The \textbf{blue curve} shows the F-Score of the goal-directed TRPO policy across different iterations. The shaded regions indicate the standard deviations across the 20 GR tasks.}
    \caption{F-Score for PandaReach environment with 10\% observability rate, evaluated across 20 different GR problems. The \textbf{orange curve} represents the F-Score of the GC-TRPO policy (for GC-AURA) across different iterations. The \textbf{blue curve} shows the F-Score of the goal-directed TRPO policy (for DRACO) across different iterations. The shaded regions indicate the standard deviations across the 20 GR problems.}
    \label{fig:gc_aura_rec}
\end{minipage}
\end{figure}

\begin{table*}[t]
\centering
\begin{tabular}{llrrrr}
    \toprule
    \textbf{OBS} & \textbf{Problem} & \textbf{Meta-AURA (100 iters)} & \textbf{DRACO (350 iters)} & \textbf{GRAQL (700 iters)} & \textbf{Ramirez \& Geffner} \\
    \midrule
    10\% & 2-Goals & 0.90 ± 0.31 & \textbf{1.00 ± 0.00} & 0.75 ± 0.32 & 0.67 ± 0.00 \\
         & 3-Goals & \textbf{0.90 ± 0.31} & \textbf{0.90 ± 0.31} & 0.64 ± 0.35 & 0.52 ± 0.00 \\
    \midrule
    30\% & 2-Goals & \textbf{1.00 ± 0.00} & \textbf{1.00 ± 0.00} & \textbf{1.00 ± 0.00} & 0.83 ± 0.48 \\
         & 3-Goals & \textbf{1.00 ± 0.00} & \textbf{1.00 ± 0.00} & \textbf{1.00 ± 0.00} & 0.77 ± 0.48 \\
    \bottomrule
\end{tabular}
\caption{Performance in MiniGrid over 20 GR settings under 10\% and 30\% partial observability (using KL-divergence as the GR distance metric). The table reports the average F-Score (± standard deviation) of the algorithms across different GR problems.}
\label{tab:minigrid}
\end{table*}

\begin{table*}[t]
\centering
\begin{tabular}{llrrrrrrrr}
    \toprule
    \textbf{OBS} & \textbf{G} & \multicolumn{4}{c}{\textbf{Meta-AURA (500 iterations)}} & \multicolumn{4}{c}{\textbf{DRACO (1000 iterations)}} \\
    \cmidrule(lr){3-6} \cmidrule(lr){7-10}
    & & \textbf{Accuracy} & \textbf{Precision} & \textbf{Recall} & \textbf{F-Score} & \textbf{Accuracy} & \textbf{Precision} & \textbf{Recall} & \textbf{F-Score} \\
    \midrule
    1\% & 2 & \textbf{0.90 ± 0.32} & \textbf{0.90 ± 0.32} & \textbf{0.90 ± 0.32} & \textbf{0.90 ± 0.32} & 0.40 ± 0.52 & 0.40 ± 0.52 & 0.40 ± 0.52 & 0.40 ± 0.52 \\
        & 3 & \textbf{0.99 ± 0.21} & \textbf{0.90 ± 0.32} & \textbf{0.90 ± 0.32} & \textbf{0.90 ± 0.32} & 0.60 ± 0.34 & 0.40 ± 0.52 & 0.40 ± 0.52 & 0.40 ± 0.52 \\
    \midrule
    3\% & 2 & \textbf{1.00 ± 0.00} & \textbf{1.00 ± 0.00} & \textbf{1.00 ± 0.00} & \textbf{1.00 ± 0.00} & 0.90 ± 0.32 & 0.90 ± 0.32 & 0.90 ± 0.32 & 0.90 ± 0.32 \\
        & 3 & \textbf{1.00 ± 0.00} & \textbf{1.00 ± 0.00} & \textbf{1.00 ± 0.00} & \textbf{1.00 ± 0.00} & 0.93 ± 0.21 & 0.90 ± 0.32 & 0.90 ± 0.32 & 0.90 ± 0.32 \\
    \bottomrule
\end{tabular}
\caption{
    Performance in PointMaze over 20 GR settings under 1\% and 3\% partial observability (OBS) using Wasserstein GR distance metric in domains with 2 or 3 goals (G). The table reports the average performance (± standard deviation) for Meta-AURA and DRACO across different GR problems.
}
\label{tab:pointmaze}
\end{table*}

\section{Results}
We evaluate AURA across two core generalization constructs: adaptation to new goals, and adaptation to new domains. The latter is further split into adaptation in discrete vs. in continuous domains. In Section \ref{sec:gc_aura_exp}, we test goal generalization in a fixed domain, comparing GC-AURA to DRACO in continuous goal spaces without additional fine-tuning. Section \ref{sec:meta_discrete} examines domain generalization in discrete environments, using Meta-AURA to adapt across changing transitions and goals in MiniGrid. Section \ref{sec:meta_continuous} extends this evaluation to continuous domains, assessing Meta-AURA’s ability to generalize across structurally varied PointMaze environments. Together, these experiments test AURA's capacity to learn transferable policies, recognize goals under uncertainty, and adapt across diverse settings.
%%%%%%%%%%%%%%%%%%%%%%%%%%%%%%
% Experiment 1: GC-AURA
%%%%%%%%%%%%%%%%%%%%%%%%%%%%%%
\subsection{Adaptation to New Goals - \textcolor{RoyalPurple}{DG} (Fixed Domains)}
\label{sec:gc_aura_exp}
% \subsubsection{Experimental Setup}
To evaluate AURA's generalizability to new goals, we compare GC-AURA to DRACO in the PandaReach environment with dense rewards. GC-AURA policies, did not require any additional fine-tuning after the initial training, while DRACO is trained from scratch for each new goal introduced.  For Goal-Directed training (DRACO), we selected three settings with specific goals. %, as shown in Table~\ref{tab:domains}. 
During the Goal-Conditioned training process, the goal locations were chosen randomly from the continuous state space in each episode. TRPO was used to train all policies. Each instance consisted of 4 possible goals, and was tested with 3 different seeds. 
For recognition performance, 20 different GR problems were evaluated. In addition, we used observation sequences with 10\% observability and various noise levels (0-90\%). $x\%$ noise implies that for each action of the agent that generated this observation sequence, with a probability of $\frac{x}{100}$, the action was chosen based on the trained policy, and with a probability of $1 - \frac{x}{100}$, the action was chosen randomly.
%of the time, the policy acted randomly, and . %The Wasserstein distance metric, as introduced in Nageris \textit{et al}. \cite{nageris2024goal}, was used to compute the distance between observations and the policy.

% \subsubsection{Results}
In terms of \textbf{learning}, as shown in Figure \ref{fig:gc_aura_train}, DRACO reached an approximate optimal policy within $\approx 400$ iterations, while GC-AURA required $\approx 1400$ iterations (approximately 3.5 times slower). However, GC-AURA trained a policy applicable across all goals in a continuous goal space, whereas the DRACO was limited to individual goals.

 In terms of \textbf{recognition quality}, Figure \ref{fig:gc_aura_rec} highlights that the GC policy converged to a perfect F-Score during recognition after 600 iterations, demonstrating robust performance even with 10\% partial observability and high noise. In contrast, despite training for 400 iterations per new goal, DRACO policies struggled to achieve such performance under the same conditions. %\matan{You say that TRPO reaching recognition quality, and that TRPO is measured on GR tasks. But TRPO is an RL algorithm, not GR. Maybe you meant GC-AURA with GC-TRPO, and DRACO with TRPO? I think it's worth using the full names since it is not clear what s measured against what.}
Thus, while GC-AURA requires longer initial training, it has two key advantages: (1) it can adapt to new goals in a continuous space without requiring additional fine-tuning, and (2) it exhibits less noise-susceptibility than DRACO's goal-directed policies.

\subsection{Adaptation to New Discrete Domains - \textcolor{orange}{D} (Changing Transitions and Goals)}
\label{sec:meta_discrete}
To evaluate AURA’s ability to generalize across new discrete settings, we compare Meta-AURA, GRAQL, DRACO, and the Ramirez \& Geffner planner-based method (R\&G). We tested each approach in three different goal configurations (Table \ref{tab:domains}) on the MiniGrid Empty-9×9 domain. Note that each method relies on a distinct input format: R\&G operates over a hand-crafted PDDL domain and problem file; GRAQL uses a symbolic MDP representation of the grid (discrete states and actions); DRACO and Meta-AURA consume raw image observations (pixel vectors) as input, without any additional symbolic context.

\begin{figure}[t]
\begin{center}
  % Top row: MiniGrid plots
  \begin{subfigure}[b]{0.45\columnwidth}
         \centering
        \includegraphics[width=\textwidth]{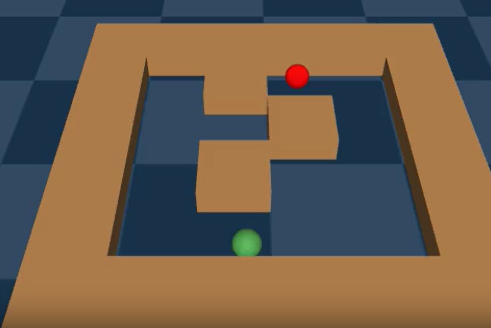}
  \caption{Training maze example 1}
        \label{fig:point_maze_train1}
  \end{subfigure}
  \hfill
  \begin{subfigure}[b]{0.45\columnwidth}
         \centering
         \includegraphics[width=\textwidth]{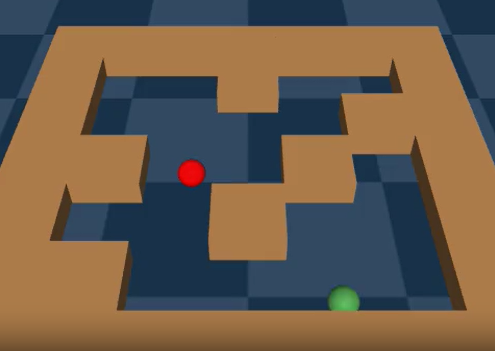}
  \caption{Training maze example 2}
        \label{fig:point_maze_train2}
  \end{subfigure}

  % Bottom row: Point Maze and Panda-Gym plots
  % \vspace{0.3cm}
  \begin{subfigure}[b]{0.45\columnwidth}
         \centering
         \includegraphics[width=\textwidth]{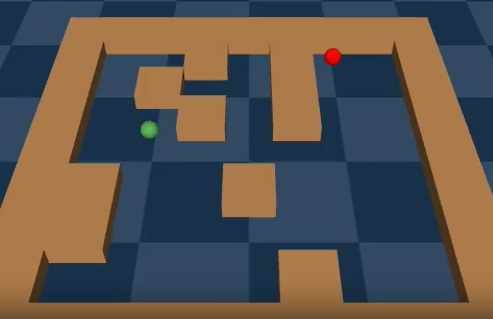}
  \caption{Training maze example 3}
        \label{fig:point_maze_train3}
  \end{subfigure}
  \hfill
  \begin{subfigure}[b]{0.45\columnwidth}
         \centering
         \includegraphics[width=\textwidth]{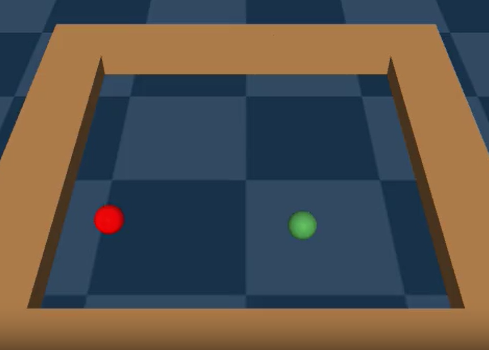}
  \caption{Training maze example 4}
        \label{fig:point_maze_train4}
  \end{subfigure}

  \caption{Examples from the train set of Point-Maze environments for MAML-RL training}
  \label{fig:point_maze_train_set}
\end{center}
\end{figure}

% We compare AURA to DRACO trained using TRPO, using a similar protocol as the previous experiment. %For Goal-Directed training (GD-AURA), we selected three environments with specific goals, as shown in Figure~\ref{fig:3goal}, and trained a policy for each goal using TRPO with multiple seeds. 
For Meta-AURA, the MAML-TRPO training involved randomly changing the goal location after each episode to vary the rewards, and placing 0–4 lava tiles in different locations within the domain to vary the transition dynamics. The meta-policy was trained using MAML-TRPO with 3 different seeds. The trained meta-policy (iteration 150 of MAML-TRPO) 
% \reuth{Did the training of the meta policy take only 150 iterations, or was that just the refinement of the meta-policy for a specific goal?} 
was then fine-tuned using TRPO for every goal, while GRAQL and DRACO policies were trained from scratch for every goal. % as the initial policy for the same three goals used in GD-AURA.

GR performance was evaluated under low observability rates, where only 10\% and 30\% of the sequence was observed. %partial observabili. 
%using four metrics (accuracy, precision, recall, and F-Score) 
Ten GR problems were evaluated for two-goal and ten for three-goal scenarios across DRACO policies (trained for $\approx 350$ iterations), Meta-AURA policies (trained for $\approx 100$ iterations), GRAQL policies (trained for $\approx 700$ iterations), and R\&G. %KL-divergence was used as the distance metric for all GR tasks.

Figure \ref{fig:minigrid_adaptation} shows a comparison of Meta-AURA and DRACO in terms of goal adaptation (training) time for a novel goal in a new domain. Meta-AURA achieved high rewards after $\approx 10$ fine-tuning TRPO iterations, while DRACO required $\approx 100$ TRPO iterations to reach similar rewards for that goal. Table \ref{tab:minigrid} indicates that Meta-AURA outperformed GRAQL and R\&G under low observability (10\% and 30\%), even with a more challenging visual representation.
% \osher{Add GPU vs. CPU, representations, offline/online}
Regarding runtime, the premise of using AURA is reducing goal adaptation and inference times.
Meta-AURA's Init Memory Phase required $150$ iterations of MAML-TRPO training, and the goal adaptation time was $\approx 100$ fine-tuning TRPO iterations. DRACO's goal adaptation time was approximately $\approx 350$ TRPO iterations, meaning that DRACO required $\approx 3.5$ times more iterations to achieve similar or higher recognition rates. Domain adaptation times are constant, as they are not used in Meta-AURA and DRACO. R\&G did not require training time but had significantly higher inference times ($\approx 2$ seconds) compared to Meta-AURA and DRACO (both with $\approx 0.13$ seconds) and GRAQL (approximately $\approx 0.1$ seconds).

% Meta-AURA's Init Information State phase time was 150 iterations of MAML-TRPO training, and the goal adaptation time was $\approx 100$ fine-tuning TRPO iterations. DRACO's goal adaptation time was $\approx 350$ TRPO iterations. Means that in order to reach similar or higher GR performance DRACO needed using $\approx 3.5$ more iterations. Note that All domain adaptation times are constant, as they are not used in Meta-AURA and DRACO. R\&G did not require training time but had significantly higher inference times ($\approx 2$ seconds) compared to Meta-AURA, DRACO (both with $\approx 0.13$ seconds) and GRAQL ($\approx 0.1$ seconds).
% Meta-AURA's domain adaptation time was $38.06 \pm 10.75$ minutes, and goal adaptation time was $64.71 \pm 2.34$ minutes. DRACO's average goal adaptation time was $77 \pm 2.29$ minutes. %, whereas GRAQL required $1108 \pm 463$ minutes. 
% R\&G did not require training time but had significantly higher inference times ($\approx 2$ seconds) compared to Meta-AURA and DRACO ($\approx 0.12$ seconds) and GRAQL ($\approx 0.1$ seconds).

%%%%%%%%%%%%%%%%%%%%%%%%%%%%%%
% Experiment 3: META-AURA in Continuous Domains
%%%%%%%%%%%%%%%%%%%%%%%%%%%%%%
\subsection{Adaptation to New Continuous Domains (\textcolor{orange}{D}) (Changing Transitions and Goals)}
\label{sec:meta_continuous}
To evaluate AURA's generalization across new continuous settings, we used the PointMaze Four-Rooms 11x11 environment with sparse rewards (goals at 1x9, 1x6 and 4x9 locations). For DRACO, we followed a similar protocol to the previous experiment. We selected three domains with specific goals and trained a policy for each goal using TRPO with multiple seeds for $\approx 1000$ iterations. 
% For GC-AURA, the environment replaced the goal location randomly across the continuous goal space after each episode, and a single policy was trained using SAC with HER for faster sparse goal-conditioned training.
For Meta-AURA, the MAML-TRPO training used domains with varying goal locations, rewards, sizes (6x6 to 9x9) and obstacles (see figure~\ref{fig:point_maze_train_set} and more details on MAML's performance in Appendix~\ref{sec:compute}). The trained meta-policy (iteration 50) 
% \reuth{Did the training of the meta policy take only 150 iterations, or was that just the refinement of the meta-policy for a specific goal?}
was then fine-tuned using TRPO for the same three goals used in DRACO  for $\approx 500$ iterations. %Unlike Minigrid, the meta-policy here demonstrated generalization to a new domain not included in the meta-training set.
GR performance was evaluated under 1\% and 3\% partial observability levels. Ten GR problems were evaluated for two-goal and ten for three-goal scenarios using DRACO (trained for $\approx 1000$ iterations) and Meta-AURA policies (trained for $\approx 500$ iterations). %, and GC-AURA policies (trained after 1500 iterations).

\begin{figure}[t]
\centering
\begin{minipage}[t]{0.48\textwidth}
    \centering
    \includegraphics[width=\textwidth]{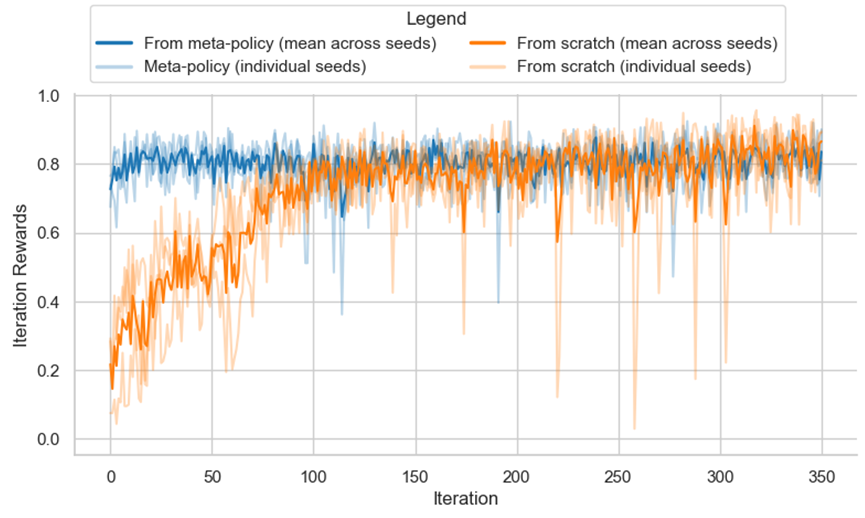}
    \caption{Adaptation to new goals (training) from a Meta-Policy and from scratch in Minigrid Empty 9x9 environment for goal 7x7.}
    \label{fig:minigrid_adaptation}
\end{minipage}
\hfill
\begin{minipage}[t]{0.48\textwidth}
    \centering
    \includegraphics[width=\textwidth]{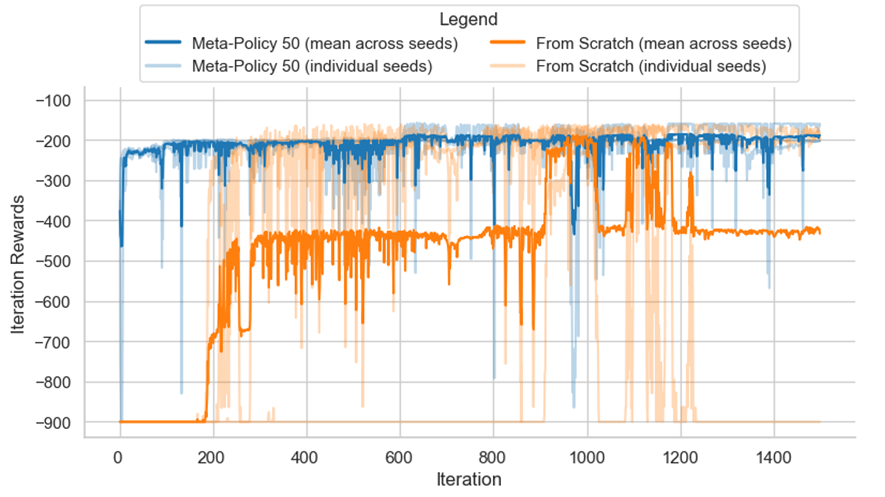}
    \caption{Adaptation to new goals (training) from a Meta-Policy and from scratch in Point-Maze 4 Rooms 11x11 with goal 1x9.}
    \label{fig:meta_aura_train}
\end{minipage}

\end{figure}

Figure \ref{fig:meta_aura_train} shows that in the goal adaptation phase, Meta-AURA policies achieved high rewards within fewer iterations ($\approx 10$), while DRACO's policies that were trained from scratch required approximately $\approx 200$ or more iterations until reaching similar rewards. %GC-AURA required 1500 iterations to train the policy.
Table \ref{tab:pointmaze} shows that the fine-tuned Meta-AURA policies (after $\approx 500$ iterations) outperformed DRACO (after $\approx 1000$ iterations) in terms of GR quality with a shorter goal adaptation phase. % while GC-AURA policies (after 1000 iterations) achieved perfect GR performance, even under 1\% partial observability. GD-AURA, despite near-optimal policies, failed to achieve perfect GR performance.

\section{Related Work}
Over the past two decades, GR has been addressed primarily through symbolic approaches, such as planning \citep{ramirez2009plan,sohrabi2016plan,meneguzzi2021survey}. These studies introduced the concept of GR by employing planners to infer the most likely goal based on given observations. %However, planning-based methods require detailed domain knowledge and often rely heavily on expert inputs, which can be impractical in dynamic or unfamiliar environments. Furthermore,
In stochastic and continuous domains GR required a shift from traditional methods toward model-free approaches. Recent advancements have explored the integration of machine learning techniques into GR problems. Notably, the GR as RL framework \citep{amado2022goal} has shown promise by leveraging RL to infer goals in such environments for discrete goal spaces, and DRACO framework \citep{fang2023real,nageris2024goal} for continuous goal spaces. Other studies reframed GR as a supervised learning problem, leveraging deep learning for goal classification \citep{min2014deep,amado2018goal,chiari2023goal}. Further approaches utilize process-mining to learn the expected processes that are likely for each goal \citep{ko2023plan,su2023fast}.
These methods are all designed for static, single GR problems, limiting their applicability in real-world settings where agents must handle continuous and changing tasks. Recent work introduced Online Dynamic Goal Recognition for grid-based domains \citep{shamir2024odgr}. This definition encompasses multiple GR tasks within the same domain and provides a proof-of-concept for dynamic GR with changing goals in a simple, discrete navigational domain. % While this work conceptualized Dynamic GR, it was limited to empty navigational domains and GR tasks within the same domain with discrete state and goal spaces. %However, these approaches depend heavily on the availability of extensive supervised datasets and often lack interpretability, which poses challenges in critical applications where understanding the rationale behind decisions is essential.
%

%Add the paragraph below if accepted:
% Recent work introduced Online Dynamic Goal Recognition \citep{shamir2024odgr}. This definition encompasses multiple GR tasks within the same domain and provides a proof-of-concept for implementing dynamic GR in a discrete, simple navigational domain. % While this work conceptualized Dynamic GR, it was limited to empty navigational domains and GR tasks within the same domain with discrete state and goal spaces.

% Fang et al. \shortcite{fang2023real} expanded the scope of GR as RL using function approximations, which enabled some generalizability to new goals. However, it primarily focused on handling multiple GR tasks within the same domain without emphasizing generalizability among different tasks.

% This research defines the GDGR problem, proposes AURA -- algorithm for GDGR, and presents a specific application of this algorithm. Preliminary results demonstrate its time efficiency in adapting to new GR tasks, highlighting its potential to advance GR in dynamic environments.

\section{Conclusion}
\label{sec:discussion}

In this paper, we introduced the \textbf{General Dynamic Goal Recognition (GDGR)} problem, a generalization of the GR problem for dynamic environments where goals and settings change over time. To address this, we proposed the \textbf{Adaptive Universal Recognition Algorithm (AURA)}, an algorithmic framework for GDGR with three abstraction levels, each of which could benefit from mechanisms that support faster adaptation, such as caching to reuse past experiences for previously seen tasks. We further provide two example implementations: Meta RL (Meta-AURA) and GCRL (GC-AURA). These algorithms were evaluated against traditional GR baselines across multiple domains to showcase AURA's ability to adapt quickly to new goals and new domains. 

Experimental results show AURA enables faster GDGR, which is essential for applications like autonomous vehicles and assistive robots. GC-AURA demonstrated scalability and adaptability to new goals without extensive retraining, while the analysis of both implementations highlighted trade-offs across abstraction levels. GC-AURA handles noise and continuous goal spaces well within a domain, whereas Meta-AURA facilitates rapid adaptation across domains. These algorithms support tailoring GR frameworks to application needs, balancing training time and inference efficiency.

\paragraph{Limitations and Assumptions.} Meta-AURA assumes GR domains share state and action spaces, varying only in transitions and rewards. GC-AURA assumes effective training across continuous goal spaces, though high-dimensional goals may require restricting the space for reliable learning. While current implementations of AURA are RL-based, AURA is defined as a general algorithm, and future work will explore symbolic or hybrid implementations to enhance generality and efficiency.

\paragraph{Future Work} We plan to further integrate Meta-RL and Goal-Conditioned RL to improve GDGR and reach real-time GR when new goals and domains are introduced on-the-fly, and extend AURA to more realistic domains. By bridging theoretical advances with practical needs, we envision AURA as a foundation for next-generation GR systems in complex, dynamic settings.

%% The file named.bst is a bibliography style file for BibTeX 0.99c
\bibliographystyle{named}
\bibliography{ijcai25}

\clearpage
\appendix

\section{Computational Resources and Efficiency}
\label{sec:compute}

This section provides details on the computational requirements of the different methods evaluated in our experiments. All experiments except for the R\&G baseline (which uses symbolic planners with their code) involve training RL policies, which require significant computational resources. In contrast, GR inference with trained RL policies is computationally light and can be executed on a CPU within seconds by simply loading the trained models.

\paragraph{Compute Settings.} All training procedures were conducted on a single Tesla K40 GPU machine, except for GRAQL and R\&G, which ran on CPU. The R\&G planner-based method completed in a few seconds per task, as it only involves symbolic planning rather than learning-based training.

\paragraph{Storage Requirements.} 
% The storage footprint varies significantly across methods:
% \begin{itemize}
%     \item \textbf{MAML-TRPO (Meta-AURA):} $\approx 150$ MB
%     \item \textbf{GC-TRPO (GC-AURA):} $\approx 40$ MB
%     \item \textbf{DRACO:} $\approx 15$ MB
%     \item \textbf{GRAQL:} $\approx 3000$ MB
% \end{itemize}
GRAQL's high storage demand stems from its reliance on tabular Q-learning, which stores a large Q-table. In contrast, DRACO, GC-TRPO and MAML-TRPO store compact neural network weights, requiring significantly less storage.

% \paragraph{Training Time.} The average training time per approach was:
% \begin{itemize}
%     \item \textbf{MAML-TRPO (Meta-AURA):} $\approx 5-8$ hours and $\approx 3-20$ minutes for the TRPO training using the MAML policy for one goal.
%     \item \textbf{GC-TRPO (GC-AURA):} $\approx 1-3$ hours
%     \item \textbf{DRACO:} $\approx 20-50$ minutes
%     \item \textbf{GRAQL:} $\approx 10$ hours
% \end{itemize}
% The large difference in training times is primarily due to two factors: (1) GRAQL is CPU-based and lacks GPU acceleration, and (2) its tabular Q-learning strategy requires intensive storage operations. In contrast, both DRACO, GC-TRPO and MAML-TRPO benefit from GPU acceleration and train compact neural models.

\begin{figure}[htbp!]
    \centering
    \includegraphics[width=1\linewidth]{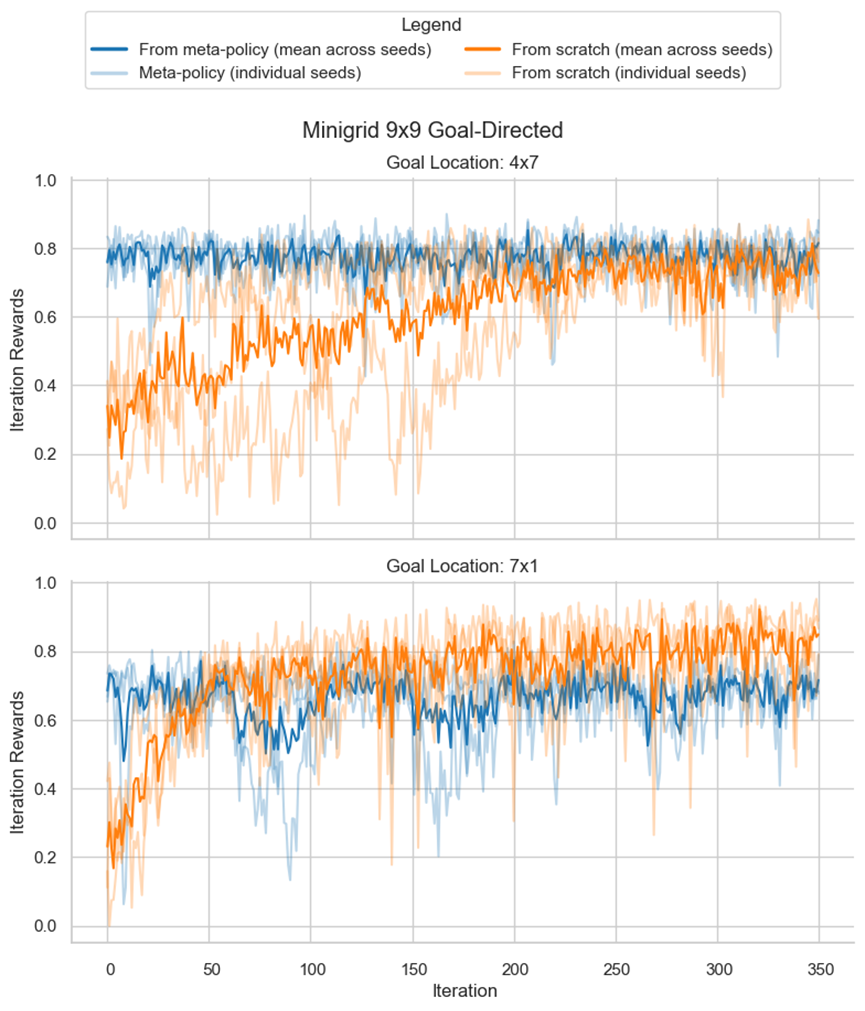}
    \caption{Adaptation to new goals (training) from a Meta-Policy and from scratch in Minigrid 9x9 Empty environment across 3 goals (7x1, 4x7).}
    \label{fig:minigrid_71_47}
\end{figure}

\begin{figure}[htbp!]
    \centering
    \includegraphics[width=1\linewidth]{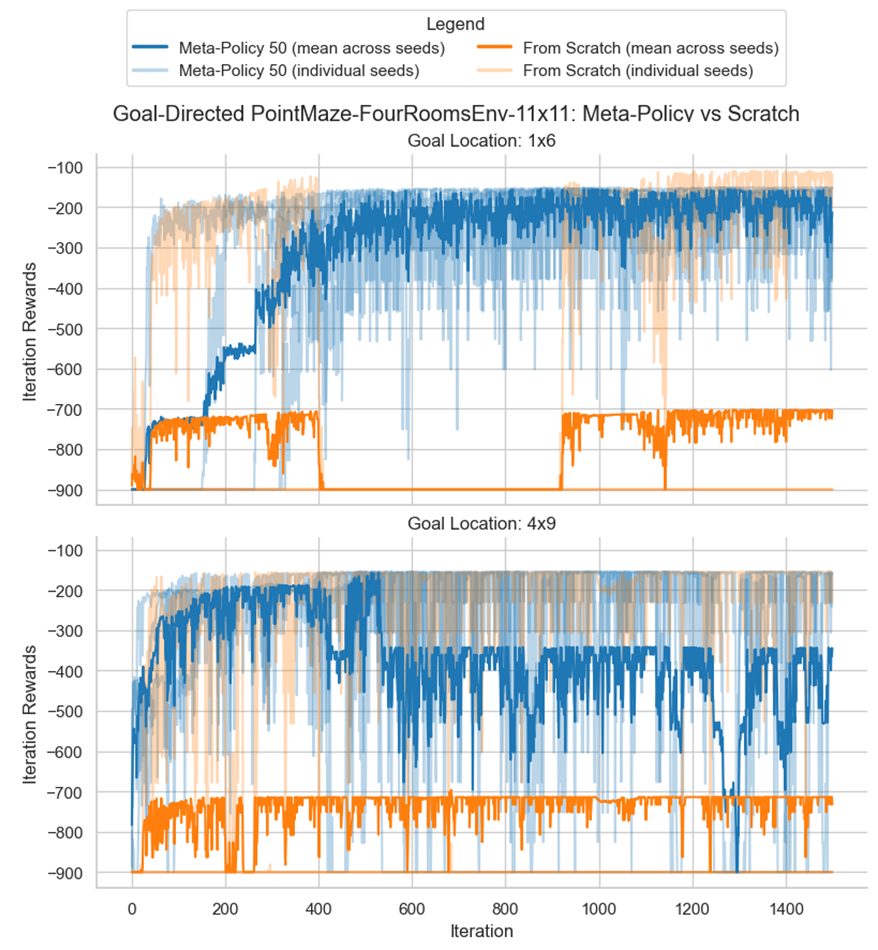}
    \caption{Adaptation to new goals (training) from a Meta-Policy and from scratch in Point-Maze 4 Rooms 11x11 environment across 2 goals (1x6, 4x9).}
    \label{fig:point_maze_16_49}
\end{figure}

\begin{figure}[htbp!]
    \centering
    \includegraphics[width=1\linewidth]{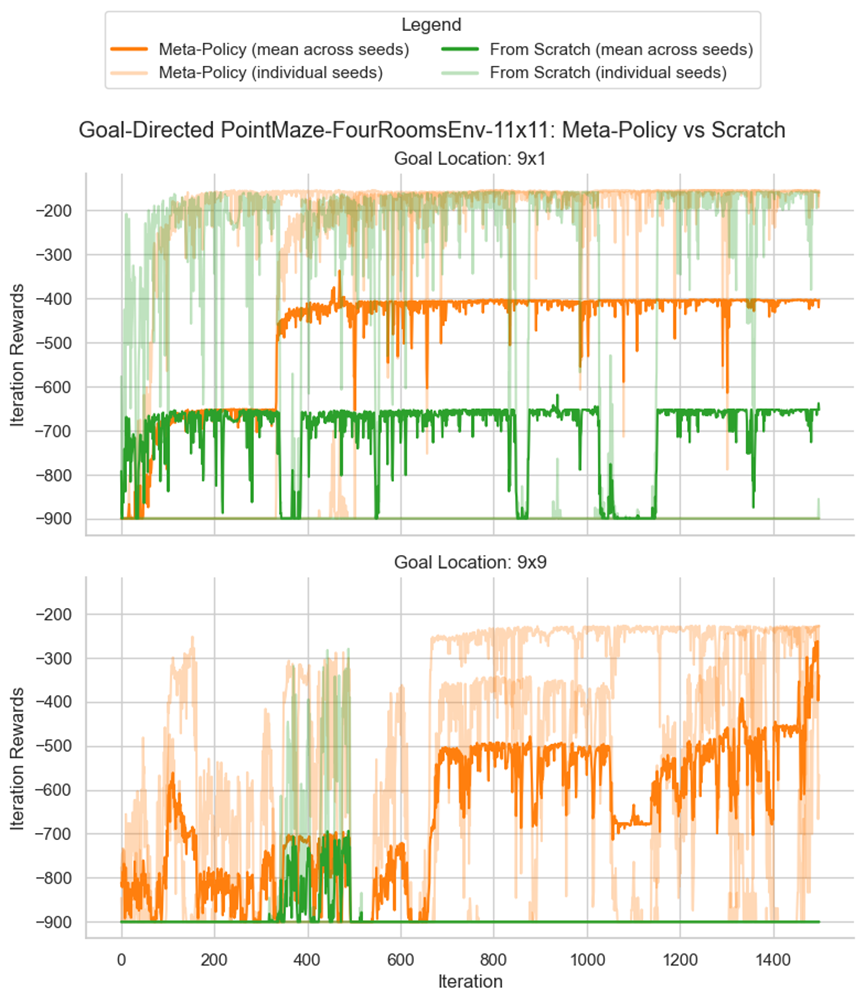}
    \caption{Adaptation to new goals (training) from a Meta-Policy and from scratch in Point-Maze 4 Rooms 11x11 environment across 2 goals (9x1, 9x9).}
    \label{fig:point_maze_91_99}
\end{figure}

\paragraph{Inference Efficiency.} After training, all GR inference tasks are fast. For RL-based methods (DRACO, GC-AURA and Meta-AURA), inference only involves a forward pass through a policy network, and thus can be executed in a few seconds even on a CPU. This makes these methods suitable for real-time applications, despite their higher training cost. Figures \ref{fig:minigrid_71_47}, \ref{fig:point_maze_16_49}, and \ref{fig:point_maze_91_99}, show the adaptation process in terms of iteration rewards, when training on new goals from scratch vs. when utilizing a meta policy. The First shows this adaptation in MiniGrid, and the latter two show this adaptation in Point-Maze settings.

\begin{figure}[t]
\begin{center}
  \includegraphics[width=0.5\textwidth]{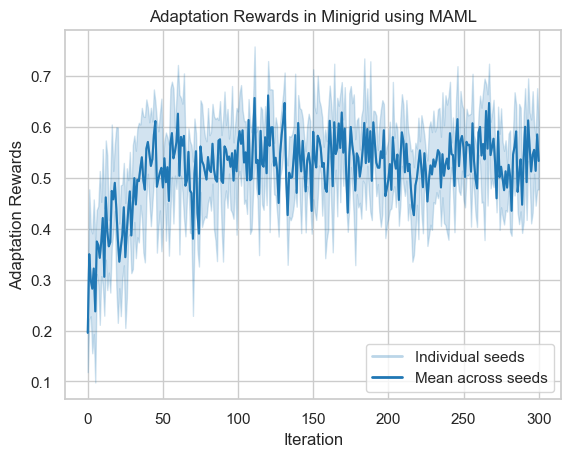}
  \caption{Minigrid MAML-TRPO training}
  \label{fig:minigrid_maml_adaptation_rewards}
\end{center}
\end{figure}

\begin{figure}[t]
\begin{center}
  \includegraphics[width=0.5\textwidth]{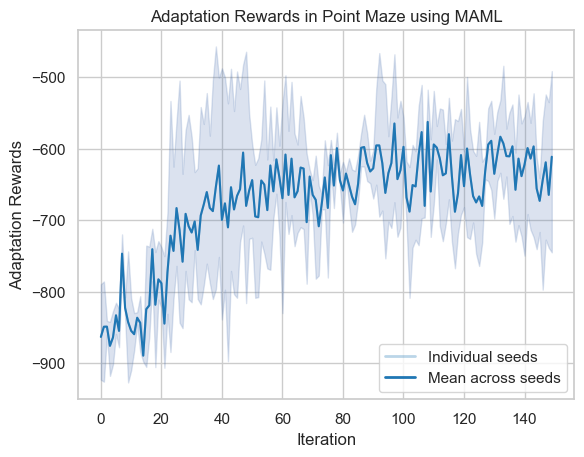}
  \caption{Point-Maze MAML-TRPO training}
  \label{fig:point_maze_maml_adaptation_rewards}
\end{center}
\end{figure}

\section{Hyperparameters}
A reasonable number of hyperparameter configurations were tested (over multiple seeds), and the results were found to be similar or less efficient than those reported here.
\label{appendix:hyperparameters}
\subsection{MAML-TRPO}
Below is an explanation of each hyperparameter used for training with MAML-TRPO:

\begin{itemize}
    \item \textbf{Adaptation batch size (\texttt{adapt\_bsz}):} The number of trajectories sampled for adaptation during the inner loop.
    \item \textbf{Adaptation learning rate (\texttt{adapt\_lr}):} The learning rate used for updating parameters during the inner loop for task-specific adaptation.
    \item \textbf{Adaptation steps (\texttt{adapt\_steps}):} The number of gradient steps performed in the inner loop for task-specific adaptation.
    \item \textbf{Backtracking factor (\texttt{backtrack\_factor}):} The factor by which the step size is reduced during line search to ensure the KL-divergence constraint is satisfied.
    \item \textbf{Discount factor (\( \gamma \)):} The discount factor for future rewards in the reinforcement learning algorithm.
    \item \textbf{Line search maximum steps (\texttt{ls\_max\_steps}):} The maximum number of steps for line search to satisfy the KL-divergence constraint.
    \item \textbf{Maximum KL divergence (\texttt{max\_kl}):} The maximum allowed KL divergence between the new and old policy distributions during updates.
    \item \textbf{Meta batch size (\texttt{meta\_bsz}):} The number of tasks sampled per iteration for meta-learning.
    \item \textbf{Meta learning rate (\texttt{meta\_lr}):} The learning rate for updating meta-parameters across tasks.
    \item \textbf{Seeds (\texttt{seed}):} Seeds used for ensuring reproducibility in experiments.
    \item \textbf{Soft update coefficient (\( \tau \)):} The coefficient for updating the policy parameters during soft updates.
\end{itemize}

Below are the hyperparameters specific to each environment:

\subsection*{MAML-TRPO Hyperparameters for MiniGrid}
\begin{itemize}
    \item Adaptation batch size (\texttt{adapt\_bsz}): 5
    \item Adaptation learning rate (\texttt{adapt\_lr}): 0.01
    \item Adaptation steps (\texttt{adapt\_steps}): 3
    \item Backtracking factor (\texttt{backtrack\_factor}): 0.5
    \item Discount factor (\( \gamma \)): 0.99
    \item Line search maximum steps (\texttt{ls\_max\_steps}): 15
    \item Maximum KL divergence (\texttt{max\_kl}): 0.01
    \item Meta batch size (\texttt{meta\_bsz}): 5
    \item Meta learning rate (\texttt{meta\_lr}): 1
    \item Seeds (\texttt{seed}): 111, 222, 333, 444
    \item Soft update coefficient (\( \tau \)): 1
\end{itemize}

\subsection*{MAML-TRPO Hyperparameters for PointMaze}
\begin{itemize}
    \item Adaptation batch size (\texttt{adapt\_bsz}): 10
    \item Adaptation learning rate (\texttt{adapt\_lr}): 0.1
    \item Adaptation steps (\texttt{adapt\_steps}): 3
    \item Backtracking factor (\texttt{backtrack\_factor}): 0.5
    \item Discount factor (\( \gamma \)): 0.95
    \item Line search maximum steps (\texttt{ls\_max\_steps}): 15
    \item Maximum KL divergence (\texttt{max\_kl}): 0.01
    \item Meta batch size (\texttt{meta\_bsz}): 10
    \item Meta learning rate (\texttt{meta\_lr}): 1
    \item Seeds (\texttt{seed}): 111, 222, 333
    \item Soft update coefficient (\( \tau \)): 1
\end{itemize}

\subsection*{Hyperparameters for TRPO Training}

Below is the set
of hyperparameters used for TRPO training across different environments.

\subsubsection*{MiniGrid (Trained from Scratch and from MAML-TRPO Policy)}
\begin{itemize}
    \item Learning rate (\texttt{lr}): 0.01
    \item Batch size (\texttt{batch\_size}): 20
    \item Discount factor (\( \gamma \)): 0.99
    \item Soft update coefficient (\( \tau \)): 1.0
    \item Random seeds (\texttt{seeds}): 100, 200, 300
    \item Maximum KL divergence (\texttt{max\_kl}): 0.01
    \item Backtracking factor (\texttt{backtrack\_factor}): 0.5
    \item Line search maximum steps (\texttt{ls\_max\_steps}): 15
\end{itemize}

\subsubsection*{PandaGym (Trained for Goal Conditioned and Goal Directed TRPO Policies)}
\begin{itemize}
    \item Learning rate (\texttt{lr}): 0.01
    \item Number of iterations (\texttt{num\_iterations}): 400
    \item Batch size (\texttt{batch\_size}): 20
    \item Discount factor (\( \gamma \)): 0.99
    \item Soft update coefficient (\( \tau \)): 1.0
    \item Random seeds (\texttt{seeds}): 100, 200, 300
    \item Maximum KL divergence (\texttt{max\_kl}): 0.01
    \item Backtracking factor (\texttt{backtrack\_factor}): 0.5
    \item Line search maximum steps (\texttt{ls\_max\_steps}): 15
\end{itemize}

\subsubsection*{PointMaze (Trained from Scratch and from MAML-TRPO Policy)}
\begin{itemize}
    \item Learning rate (\texttt{lr}): 0.001
    \item Number of iterations (\texttt{num\_iterations}): 100
    \item Batch size (\texttt{batch\_size}): 20
    \item Discount factor (\( \gamma \)): 0.99
    \item Soft update coefficient (\( \tau \)): 1.0
    \item Random seeds (\texttt{seeds}): 2, 3, 4
    \item Maximum KL divergence (\texttt{max\_kl}): 0.01
    \item Backtracking factor (\texttt{backtrack\_factor}): 0.5
    \item Line search maximum steps (\texttt{ls\_max\_steps}): 15
\end{itemize}

\subsection*{Policies for TRPO Training}

Different policies were used based on the action space of each environment:

\subsubsection*{MiniGrid (Discrete Action Space)}
For MiniGrid, we used a TRPO policy with a critic network based on a categorical policy. This network employs two hidden layers, each with 100 units, and uses ReLU activation. The policy outputs a categorical distribution over actions. The key hyperparameters are:
\begin{itemize}
    \item Input size: Dimension of the state space.
    \item Output size: Number of discrete actions.
    \item Hidden layers: [100, 100]
    \item Activation function: ReLU
\end{itemize}

\subsubsection*{PandaGym and PointMaze (Continuous Action Space)}
For PandaGym and PointMaze, we used a TRPO policy with a critic network based on a diagonal normal policy. This network includes two hidden layers, each with 100 units, and supports either ReLU or Tanh activation functions. It models actions as a normal distribution with learned mean and variance. The key hyperparameters are:
\begin{itemize}
    \item Input size: Dimension of the state space.
    \item Output size: Dimension of the continuous action space.
    \item Hidden layers: [100, 100]
    \item Activation function: ReLU
    \item Initial standard deviation (\( \sigma \)): \( \log(1) \)
\end{itemize}

\subsubsection*{Value Network for All TRPO Policies}
The value network for all TRPO policies is a linear state-value function that minimizes a least-squares loss. It uses features derived from the state, including polynomial terms and additional regularization to improve stability. The key hyperparameters are:
\begin{itemize}
    \item Input size: Dimension of the state space.
    \item Regularization coefficient (\texttt{reg}): \( 10^{-2} \)
    \item Features: Polynomial transformations of the input state.
\end{itemize}

\section{Licenses and External Packages}
\label{sec:licenses}

Our experimental framework makes use of several open-source environments and libraries. Below we provide license details, package versions, and citations for reproducibility:

\paragraph{MiniGrid} 
We used version 2.5.0 of the \texttt{minigrid} package \citep{chevalier2024minigrid}, which is licensed under the \textbf{Apache License 2.0}. This license allows for free use, modification, and distribution, provided compliance with its terms.

\paragraph{PointMaze} 
We used version 1.3.1 of the \texttt{gymnasium-robotics} package \citep{gymnasium_robotics2023github}, which provides the PointMaze environment. It is licensed under the \textbf{MIT License}, permitting unrestricted reuse in source and binary forms with proper attribution.

\paragraph{Panda-Gym} 
The Panda-Gym environment \citep{gallouedec2021panda} was used via version 3.0.7 of the \texttt{panda-gym} package. This package is distributed under the \textbf{MIT License}, allowing flexible use and distribution.

\paragraph{Learn2Learn} 
Meta-RL experiments (Meta-AURA) were implemented using version 0.2.1 of the \texttt{learn2learn} Python library \citep{Arnold2020-ss}. The library is released under the \textbf{MIT License}.

\paragraph{DRACO, GR-as-RL, and R\&G Implementations} 
The DRACO \citep{nageris2024goal} and GRAQL \citep{amado2022goal} baselines, as well as the R\&G planner-based GR algorithm \citep{ramirez2009plan}, were either reimplemented following the original papers or executed using original codebases when granted by their respective authors. Where applicable, we contacted the authors and obtained permission to use their implementations to ensure experimental accuracy and comparability.

All used tools and environments are either licensed for academic use or reproduced from cited research. These resources enable the reproducibility of our GDGR framework and its associated baselines.

\end{document}